\documentclass[runningheads]{llncs}

\usepackage{eccv}

\usepackage{eccvabbrv}

\usepackage{graphicx}
\usepackage{amsmath}
\usepackage{amssymb}
\usepackage{enumitem}
\usepackage{booktabs}
\usepackage{bbding}
\usepackage{caption}
\usepackage{soul}
\usepackage{bbm}
\usepackage{subcaption}
\usepackage{multirow}
\usepackage[accsupp]{axessibility}
\usepackage{xcolor}
\usepackage{colortbl}
\usepackage{makecell}
\usepackage[ruled,vlined]{algorithm2e} 
\usepackage[
    pagebackref,
    breaklinks,
    colorlinks=true,
    linkcolor=magenta,
    citecolor=magenta,
    urlcolor=magenta
]{hyperref}
\usepackage[accsupp]{axessibility}  % Improves PDF readability for those with disabilities.

\usepackage{hyperref}

\usepackage{orcidlink}

\begin{document}

\title{UC-VLM: Consistency-Driven Learning for AI-Generated Image Detection with \\ Vision-Language Large Models} 

\titlerunning{UC-VLM}

\author{Lei Tan\inst{1}\orcidlink{0000-0002-2766-681X} \and
Shuwei Li\inst{1}\orcidlink{0009-0002-0148-7351} \and
Mohan Kankanhalli\inst{1}\orcidlink{0000-0002-4846-2015} \and
Robby T. Tan\inst{1}\orcidlink{0000-0001-7532-6919}
}

\authorrunning{L. Tan et al.}

\institute{National University of Singapore, Singapore \\
\email{lei.tan@nus.edu.sg, shuwei@u.nus.edu, mohan@comp.nus.edu.sg, robby.tan@nus.edu.sg}}

\maketitle
\begin{abstract}
Vision-Language Large Models (VLLMs) are promising for AI-generated image (AIGI) detection because they can produce both a prediction and a natural-language output. However, most existing VLLM-based detectors primarily fine-tune the language side while giving limited attention to low-level visual forensic cues. They also often depend on manually crafted prompts or human-annotated rationales, which limits scalability.
We present \textbf{UC-VLM}, a unified multi-stage framework for AIGI detection that relies solely on binary supervision. UC-VLM first identifies effective instruction variants automatically. It then reuses the same binary label within a multi-stage training framework: (i) a visual discrimination objective that strengthens sensitivity to non-semantic forensic cues, and (ii) a label-conditioned generation objective that uses the binary label to supervise textual outputs. This design turns weak binary supervision into a shared supervision signal for both the visual pathway and the language output.
Our key novelty is a unified multi-stage binary-supervised framework that consistently reuses the same authenticity labels for visual adaptation and label-conditioned text generation, while leveraging automatically optimized instructions to reduce prompt sensitivity without requiring human-written rationales or hand-crafted prompts.
Experiments show that UC-VLM achieves \textbf{96.1\%} average accuracy on GenImage, exceeding the strongest prior result by \textbf{4.6\%}, and obtains \textbf{69.6\% / 77.9\%} accuracy on Chameleon under ProGAN / SDV1.4 training, surpassing the best baseline by \textbf{11.2\% / 15.3\%}, respectively.

\end{abstract}

\begin{figure}[t]
\centering
\includegraphics[width=\columnwidth]{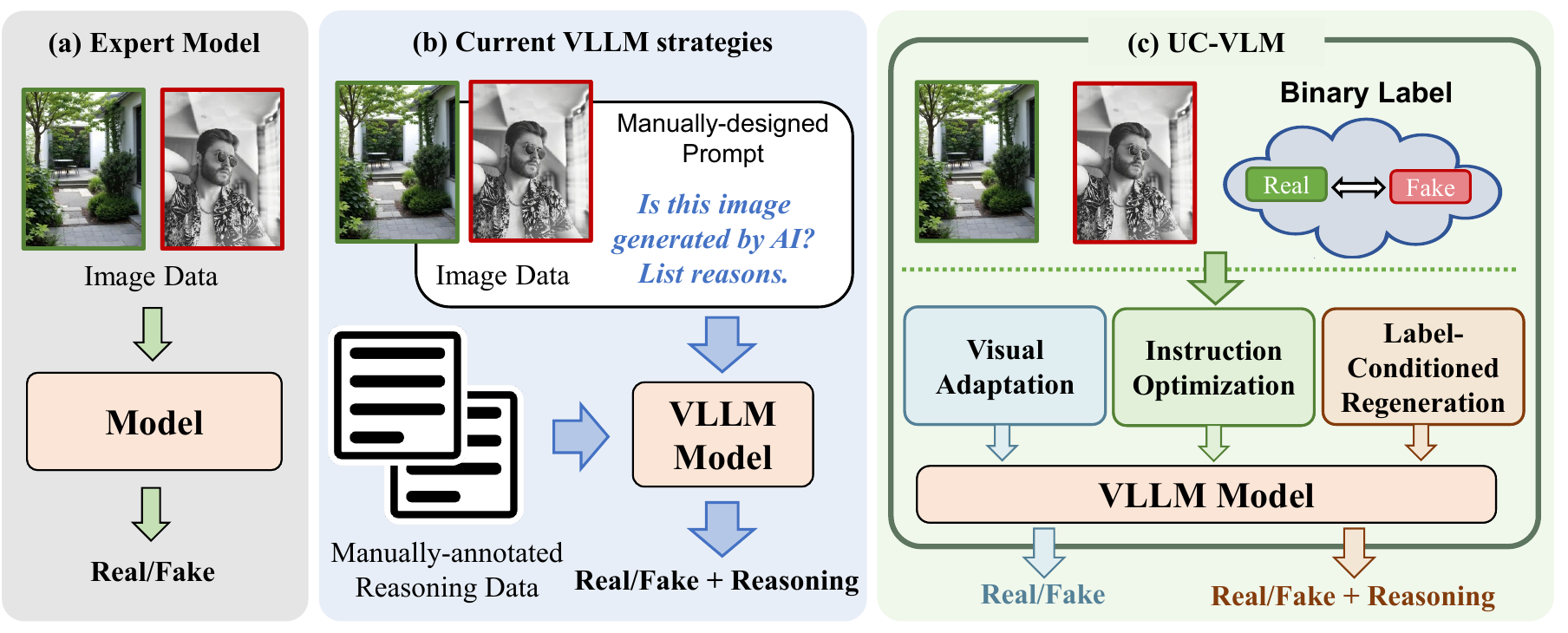}
\vspace{-1em}\caption{\textbf{Comparison of paradigms for AI-generated image detection.}
(a) Conventional classifiers predict real/fake from image data.
(b) Existing VLLM-based detectors often rely on manually crafted prompts and human-annotated rationales to produce a label with accompanying text.
(c) UC-VLM learns from binary authentic/generated labels only, combining visual adaptation, instruction optimization, and label-conditioned generation to improve robustness to prompt choice and provide scalable supervision for textual outputs.}\vspace{-2em}
\label{fig1}
\end{figure}

\section{Introduction}
\label{sec:intro}
Recent advances in generative AI have made it possible to synthesize highly realistic images that can convincingly mimic natural photographs~\cite{song2019generative,peebles2023scalable,tian2024visual,ma2026flow,li2026bridging,liu2026tele}. These capabilities are transforming content creation, but they also intensify concerns about the authenticity and trustworthiness of digital media~\cite{li2026vision,esser2024scaling,li2026vid,li2026vimu,li2026sponge}.
Most existing AI-generated image (AIGI) detectors formulate the task as binary classification~\cite{tan2024rethinking,sarkar2024shadows}. As illustrated in Figure~\ref{fig1}(a), models are trained on datasets containing both real and synthesized images and learn a discriminative boundary between the two classes. To improve separability, prior work often augments the input with explicit transformations or auxiliary signals~\cite{tan2024rethinking,wang2023dire,luo2024lare}, leveraging cues such as neighboring-pixel statistics~\cite{tan2024rethinking}, high-frequency artifacts~\cite{tan2024frequency}, or reconstruction-based discrepancies~\cite{ricker2024aeroblade}. While these strategies can yield strong benchmark accuracy, they do not naturally support textual outputs and often show limited robustness beyond their training conditions.

With recent advances in Vision-Language Large Models (VLLMs) (e.g.,~\cite{hurst2024gpt,liu2023visual,wang2025internvl3,zhang2025mamba}), research has begun to explore their potential for AIGI detection~\cite{wen2025spot,gao2025fakereasoning,zhou2025aigi}. Unlike traditional classifiers, VLLMs align visual and textual representations through large-scale multimodal pretraining, enabling a unified interface for prediction and text generation. Despite this promise, current VLLM-based detectors~\cite{wen2025spot,gao2025fakereasoning} still face key limitations, as illustrated in Figure~\ref{fig1}(b). Prior successful detectors have shown that the discriminative differences between real and generated images often lie in subtle low-level details, such as frequency artifacts~\cite{tan2024frequency}, local textures~\cite{tan2024rethinking}, and non-semantic forensic traces~\cite{cozzolino2025zero}. They often prioritize language-side fine-tuning while giving limited attention to low-level forensic cues in the visual pathway. They also rely on manually crafted prompts and human-annotated rationales, which limits scalability. Moreover, such handcrafted designs underexplore how prompt formulation affects prediction stability and detection consistency.

To address these challenges, we introduce \textbf{UC-VLM}, a unified learning framework for AIGI detection, as illustrated in Figure~\ref{fig1}(c).
UC-VLM is trained with only authentic/generated labels, but it treats these binary labels as more than final decision targets.
Our key idea is to expand weak binary supervision into a shared training signal that is consistently reused across visual adaptation, instruction optimization, and language-side generation in a unified multi-stage framework.
This shared-supervision design is important because, in VLLM-based detection, prediction stability is jointly affected by (i) what evidence the vision encoder exposes, (ii) how the instruction queries that evidence, and (iii) how the language model expresses the output under the same supervision.

Optimizing these aspects in isolation can be unreliable: visual tuning alone may overfit to low-level patterns, prompt selection alone can lead to unstable behavior under rephrasing, and language-side tuning alone may improve text fluency without improving the underlying visual evidence used for detection.
UC-VLM addresses this by reusing the same binary labels to enforce three tightly connected behaviors within a single learning framework.
First, UC-VLM adapts the vision encoder to increase sensitivity to local non-semantic forensic cues, encouraging the model to rely less on high-level semantics and more on artifact-level evidence.
Second, UC-VLM performs instruction optimization to automatically discover effective instruction formulations for authenticity assessment. 
By selecting the best-performing instruction from an automatically generated pool, the model obtains a more stable linguistic interface.
Third, UC-VLM applies label-conditioned text generation so that the binary label also provides supervision for language-side outputs, enabling scalable textual-output training without human-annotated rationales.
Together, these behaviors turn binary authenticity labels into multi-stage optimization on visual discrimination, prompt robustness, and textual-output generation, under a shared supervision signal.

Concretely, UC-VLM is implemented as a unified multi-stage training procedure. It first adapts the vision encoder with a visual discrimination loss, and then refines the language module with a label-conditioned generation loss while keeping the vision tower fixed. Although these stages are optimized sequentially, they are unified by the same binary authenticity supervision: the visual stage shapes the evidence available to the model, while the generation stage transfers the same supervision to language-side outputs. In this sense, UC-VLM is not a set of independently supervised modules, but a multi-stage binary-supervised framework that reuses the same labels across visual and language components.
Overall, our contributions are summarized as follows:
\begin{itemize}
\item We propose \textbf{UC-VLM}, a unified learning framework for VLLM-based AI-generated image detection using only binary authentic/generated labels. UC-VLM combines visual adaptation, instruction optimization, and label-conditioned generation within a unified multi-stage framework, enabling optimization of the visual pathway and language-side outputs under the same supervision signal.
\item We show that unified binary-supervised learning improves detection stability beyond standard accuracy. UC-VLM automatically discovers effective instruction formulations and enables scalable textual-output training under binary supervision, without manual prompts and rationales.
\item We conduct extensive experiments on GenImage and Chameleon, showing that UC-VLM improves accuracy over the prior results by \textbf{+4.6\%} on GenImage and \textbf{+11.2\% / +15.3\%} on Chameleon under ProGAN / SDV1.4 training, and yields more stable predictions under instruction variants.
\end{itemize}

\section{Related Work}
AI-generated image detection has become increasingly important with the rapid progress of modern generative models. Existing methods can be broadly grouped into three lines: hand-crafted forensic cues, deep-learning-based visual detectors, and recent VLM-based detectors.

Early forensic approaches mainly relied on manually designed low-level cues, such as chromatic aberration~\cite{mayer2018accurate}, color saturation~\cite{mccloskey2019detecting}, blending boundaries~\cite{li2020face}, and reflection inconsistencies~\cite{o2012exposing}. These methods provide interpretable evidence for specific types of manipulation or synthesis artifacts. However, as generative models become increasingly diverse and photorealistic, such hand-crafted cues are often brittle and may fail to generalize across different generators, resolutions, or post-processing conditions.

With the success of deep learning~\cite{he2016deepresnet,radford2021learningclip,ma2023ompq,ma2023solving,ma2024outlier,zhao2026just,cao20263dot,cao2026smart}, AIGI detectors~\cite{wang2020cnnspot,liu2020global,zhang2019detecting,liu2021swin} have achieved strong in-distribution performance. CNNSpot~\cite{wang2020cnnspot} shows that standard convolutional architectures can capture discriminative synthesis traces, while FreDect~\cite{frank2020leveraging} exploits frequency-domain representations to reveal artifacts that are less visible in the spatial domain. As diffusion models improve perceptual quality and weaken obvious low-level cues, recent methods have explored subtler discriminative cues to improve generalization~\cite{wu2023generalizable,zhang2023adaptive}. DIRE~\cite{wang2023dire} leverages reconstruction discrepancies to distinguish real and generated images, while localized and region-level representations have also been explored to capture fine-grained visual structures, which have been further adopted by other works~\cite{luo2024lare,zhang2024heap}. ZED~\cite{cozzolino2025zero} identifies discrepancies between real and generated images via coding differences from a lossless encoder in a multi-resolution structure. FatFormer~\cite{liu2024forgery}, Effort~\cite{yan2025orthogonal}, and MCAN~\cite{tan2026aggregating} improve CLIP-based detectors through frequency-domain fine-tuning, orthogonal-space adaptation, and multi-cue aggregation, respectively. NPR~\cite{tan2024rethinking} revisits upsampling artifacts to amplify subtle synthesis cues, while AIDE~\cite{yan2025sanity} and DEUA~\cite{huang2025diffusion} enhance detection with multi-frequency learning and uncertainty estimation. Despite these advances, most deep detectors remain purely visual. They usually treat binary labels only as classification targets and provide limited language-side reasoning, making their decisions less transparent and less flexible for VLM-based detection scenarios.

Recently, large Vision-Language Models (VLMs) have motivated a new direction~\cite{wang2025gpt,li2026white} for AIGI detection by enabling language-conditioned prediction and textual responses. FakeVLM~\cite{wen2025spot} fine-tunes the language component of a VLM on the well-annotated FakeClue dataset to produce authenticity judgments with accompanying text. FakeReasoning~\cite{gao2025fakereasoning} employs dual visual encoders by combining CLIP and DINO features, providing complementary visual cues for language-side generation. AIGI-Holmes~\cite{zhou2025aigi} integrates CLIP and NPR representations and adopts human-aligned Direct Preference Optimization to improve output consistency. These methods demonstrate the promise of VLM-based detectors, but they often rely on manually curated VQA-style prompts, human-written explanations, or annotated rationales. Such requirements increase annotation cost and may limit scalability when moving to new datasets or generators.

Different from these prior directions, UC-VLM aims to learn a VLM-based detector using only binary supervision. Instead of requiring human-written rationales, UC-VLM consistently reuses the same real/fake labels across visual adaptation, instruction optimization, and label-conditioned language refinement. This design enables binary supervision to guide both the visual pathway and language-side output generation in a scalable framework.

\begin{figure}[t]
\centering
\includegraphics[width=\columnwidth]{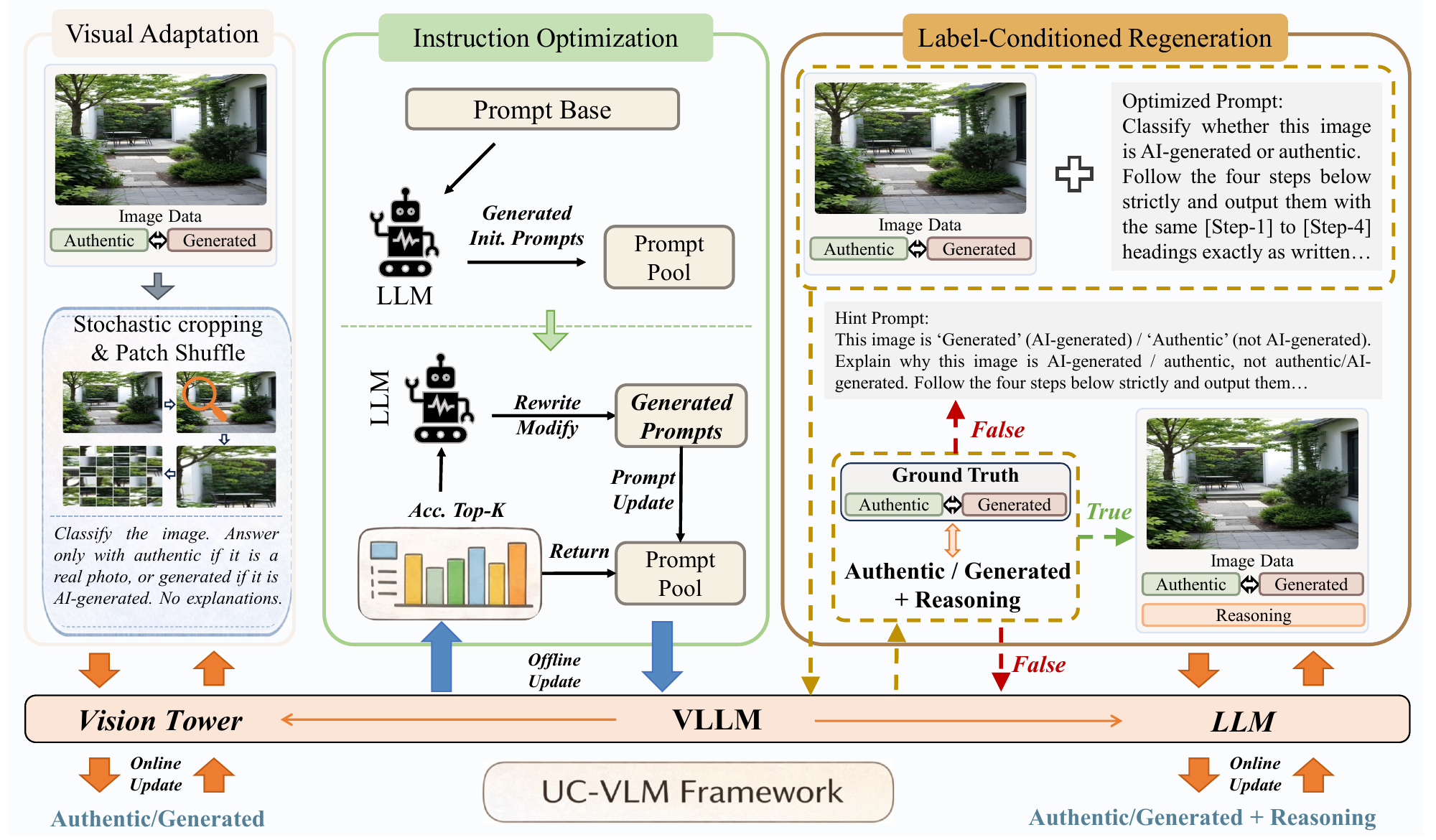}
\vspace{-1em}\caption{\label{fig_overall}\textbf{Overview of UC-VLM.}
UC-VLM is built upon Qwen2.5-VL-7B and learns AI-generated image detection under binary authentic/generated supervision.
The framework includes three coordinated components:
(i) \textbf{Visual Adaptation}, which enhances the vision encoder’s sensitivity to local non-semantic forensic cues;
(ii) \textbf{Instruction Optimization}, which automatically searches for effective instruction variants for authenticity assessment; and
(iii) \textbf{Label-Conditioned Regeneration}, which conditions text generation on the binary label to provide scalable supervision for textual outputs while freezing the vision tower.}\vspace{-2em}
\end{figure}

\section{Proposed Method}
The overall framework of UC-VLM is illustrated in Figure~\ref{fig_overall}.
Built upon a vision-language backbone (Qwen2.5-VL-7B)~\cite{bai2025qwen2}, UC-VLM addresses AIGI detection under binary authenticity supervision.
Given an input image $I$, the objective is to predict its authenticity label $y \in \{\text{Authentic}, \text{Generated}\}$, indicating whether the image is captured from the real world or synthesized by a generative model.
During training, only image-level binary labels are available, without human-written rationales, manipulation masks, or fine-grained forensic annotations.
Given an image-label pair $(I,y)$, UC-VLM reuses the binary authenticity label as the common supervision source across multiple training stages.
Rather than using the binary label only as a final classification target, UC-VLM expands this weak supervision into three complementary aspects of learning.
Specifically, the framework includes (i) a visual adaptation stage that strengthens the vision encoder’s sensitivity to non-semantic forensic cues, (ii) an instruction optimization component that automatically identifies an effective instruction formulation for authenticity assessment, and (iii) a label-conditioned generation stage that transfers the same binary supervision signal to language-side outputs while freezing the vision tower.
In this way, non-semantic forensic cues, prompt robustness, and textual-output training are consistently guided by the same binary label, rather than requiring extra sources of supervision.
The training objectives used across the stages of UC-VLM can be summarized as:
\begin{equation}
\mathcal{L}=\mathcal{L}{_{vis}}+\mathcal{L}{_{text}},
\end{equation}
where $\mathcal{L}{_{vis}}$ is used for visual-pathway adaptation and $\mathcal{L}{_{text}}$ is used for label-conditioned text generation.
In practice, these objectives are instantiated in a multi-stage training procedure, with visual adaptation performed before language-side generation.
Thus, UC-VLM provides a unified binary-supervised framework by consistently reusing the same authenticity labels across visual adaptation, instruction optimization, and language-side generation, rather than introducing additional rationale annotations.

\subsection{Visual Adaptation}
Prior studies on AIGI detection~\cite{tan2024rethinking,luo2024lare} show that authenticity discrimination often depends on subtle low-level artifacts, such as texture inconsistencies, boundary distortions, and frequency irregularities.
However, the visual backbone of a VLLM is typically optimized for general semantic understanding, which may underemphasize the local non-semantic evidence needed for forensic discrimination.
To improve sensitivity to such cues, we introduce a visual adaptation mechanism that suppresses reliance on global semantics while preserving local structural statistics.

Specifically, given an input image $I \in \mathbb{R}^{H \times W \times 3}$, we first sample a local crop $I' \in \mathbb{R}^{h' \times w' \times 3}$, where
\begin{equation}
h' \sim U\left(\tfrac{1}{8}H, \tfrac{1}{4}H\right), \quad 
w' \sim U\left(\tfrac{1}{8}W, \tfrac{1}{4}W\right).
\end{equation}
This stochastic cropping limits access to full-scene semantics while retaining fine-grained textures.
As a result, the model is encouraged to rely less on object identity or scene composition and more on local evidence relevant to authenticity.

Following the patch-based processing of Qwen2.5-VL-7B, the cropped image is divided into $N$ non-overlapping patches $\{p_i\}_{i=1}^{N}$.
We further randomly permute their spatial order:
\begin{equation}
\tilde{I}' = \text{Reassemble}\left(\{p_{\pi(i)}\}_{i=1}^{N}\right), \quad \pi \sim \mathcal{P}(N).
\end{equation}
This patch shuffling disrupts global spatial coherence while preserving local statistics.
Consequently, the model is discouraged from depending on high-level semantic layout and instead pushed toward textural and structural irregularities that are more relevant for distinguishing authentic from generated images.

The transformed image is then encoded into visual embeddings $\{\mathbf{v}_i\}_{i=1}^N$, which are fused with textual tokens to produce a binary authenticity prediction:
\begin{equation}
\hat{y} = f_{\text{VLM}}\!\left(\{\mathbf{v}_i\}_{i=1}^N, \mathbf{t}\right).
\end{equation}
We optimize the visual adaptation term using binary cross-entropy,
\begin{equation}
\mathcal{L}_{\text{vis}} = 
- \left[y \log(\hat{y}) + (1 - y) \log(1 - \hat{y})\right].
\end{equation}
Under this objective, the vision pathway is directly tuned toward artifact-aware representations using the same authenticity supervision signal that also drives the rest of UC-VLM.
This design is important for the overall framework: the visual term does not act as a standalone pretraining step, but as one component of the shared binary-supervised learning process that supports instruction robustness and language-side refinement.

\begin{table}[t]
\centering
\caption{\textbf{Effect of prompt formulation on the Chameleon dataset using the pretrained Qwen2.5-VL-7B model.}
Structured prompts (e.g., P-3) yield higher detection accuracy, indicating that VLM performance is strongly influenced by prompt formulation.}
\label{tab:prompt_vari}
\resizebox{0.5\columnwidth}{!}{\begin{tabular}{l|c}
\toprule
\textbf{Prompt ID} & \textbf{Accuracy (\%)} \\ 
\midrule
P-1: \textit{Authentic/Generated QA} & 55.5 \\
P-2: \textit{Real/Fake QA} & 59.8 \\
P-3: \textit{Answer with Explanation} & 63.6 \\
\bottomrule
\end{tabular}}\vspace{-1em}
\end{table}

\begin{algorithm}[t]
\caption{Instruction Optimization}
\label{alg:instruction_optimization}
\DontPrintSemicolon
\KwIn{LLM, VLM, BASE\_PROMPT, pool size $N$, Top-$k$ ($k<N$), rounds $M$}
\KwOut{Prompt-B, Instruction Pool}

Initialize Instruction Pool $\mathcal{P}_0=\{p_i\}_{i=1}^{N}$ generated by LLM from BASE\_PROMPT\;

\For{$t=0$ \KwTo $M-1$}{
Evaluate each $p\in\mathcal{P}_t$ on VLM, obtain $\text{Acc}(p)$\;
Select $\mathcal{P}^*_t = \text{Top-}k(\mathcal{P}_t)$ by $\text{Acc}(p)$, set Prompt-B $=\arg\max_{p\in\mathcal{P}^*_t}\text{Acc}(p)$\;
Sample Random Parent Instructions $\mathcal{R}_t \subseteq \mathcal{P}^*_t$\;
Generate Refined Instructions $\tilde{\mathcal{P}}_t$ using LLM: \\
\Indp
$p' = 
\begin{cases}
\text{Rewrite}(p), & \text{with prob. } 0.5\\
\text{Modify}(p), & \text{with prob. } 0.5
\end{cases}$, $\forall p \in \mathcal{R}_t$
\Indm
\;
Update Instruction Pool: $\mathcal{P}_{t+1} = \mathcal{P}^*_t \cup \tilde{\mathcal{P}}_t$, $|\mathcal{P}_{t+1}|=N$\;
}
\Return Prompt-B, Instruction Pool
\end{algorithm}
\subsection{Instruction Optimization}
The formulation of instructions can significantly influence the behavior of VLLMs.
Here, we use the term \emph{instruction} to denote the textual query used for authenticity assessment.
To illustrate this sensitivity, we evaluate the backbone model (Qwen2.5-VL-7B) on the Chameleon dataset under different instruction formulations:

\begin{quote} \small
\noindent \textbf{P-1}: {\ttfamily Classify if this image is AI-generated or authentic. Answer only with ``Authentic'' or ``Generated''. No explanations.}

\vspace{0.2cm}
\noindent \textbf{P-2}: {\ttfamily Classify if this image is real or fake. Answer only with ``Real'' or ``Fake''. No explanations.}

\vspace{0.2cm}
\noindent \textbf{P-3}: {\ttfamily Evaluate the image and decide if it is ``Generated'' (AI-generated) or ``Authentic'' (not AI-generated). Follow the four steps below.}
\begin{itemize}
    \item {\ttfamily Step-1: Visual Examination: List up to three suspicious and three realistic visual cues (e.g., glitches, lighting, motion blur).}
    \item {\ttfamily Step-2: Statistical Analysis: Highlight frequency-domain evidence supporting or contradicting authenticity.}
    \item {\ttfamily Step-3: Semantic \& Physics Consistency: Identify impossible \\ reflections, shadow errors, or physical mismatches.}
    \item {\ttfamily Step-4: Final Assessment: Give one word stating ``Generated'' or \\ ``Authentic'', and summarize key evidence.}
\end{itemize}
\end{quote}

As shown in Table~\ref{tab:prompt_vari}, even minor lexical variations can noticeably affect detection accuracy.
Replacing `Authentic/Generated' with `Real/Fake' already yields a measurable improvement, while adding structured guidance further improves performance.
These results indicate that the backbone VLLM already contains useful knowledge for authenticity discrimination, but its predictions remain sensitive to the linguistic form of the instruction.

Motivated by this sensitivity, we introduce an automatic instruction optimization procedure to identify instruction variants that are both effective and relatively stable for AIGI detection.
Rather than relying on manually designed prompts, we construct an instruction pool and iteratively refine it through generation, evaluation, and transformation, as illustrated in Figure~\ref{fig_overall} and Algorithm~\ref{alg:instruction_optimization}.

Starting from a structured base template (\textit{BASE\_PROMPT}), the LLM generates $N$ candidate instructions to initialize an instruction pool.
Each candidate is evaluated on a validation subset using the VLM, and its detection accuracy is recorded.
The top-$k$ high-performing instructions are retained, and the best-performing instruction among them is denoted as \textit{Prompt-B}.

To encourage linguistic diversity while preserving semantic intent, a subset of the retained instructions is randomly selected as parent instructions and sent back to the LLM for transformation.
For each selected parent instruction, the LLM performs either semantic rewriting, which preserves meaning while changing wording, or reasoning-expression modification, which changes the structural form of the instruction, each with probability 0.5.
This process generates a set of refined child instructions.

The refined instructions are merged with the retained top-$k$ parents to form the updated instruction pool while maintaining a fixed pool size of $N$.
The generation-evaluation-refinement process is repeated for $M$ rounds, gradually exploring diverse yet semantically related instruction formulations.
At the end of optimization, the resulting pool provides a family of instruction variants for authenticity assessment, and the best-performing instruction \textit{Prompt-B} is used as the default query in the following experiments.

\subsection{Label-Conditioned Regeneration}
Under binary authenticity supervision, VLLM training directly supervises the final authenticity prediction, but does not provide explicit supervision on the content of the generated text.
To reuse the same binary supervision signal on the language side, without collecting human-written rationales, we introduce a label-conditioned regeneration mechanism.

Using the optimized instruction $p_{\text{best}}$, the VLM first produces a textual output and an authenticity prediction for each image $x_j$:
\begin{equation}
(r_j, \hat{y}_j) = f_{\text{VLM}}(x_j, p_{\text{best}}),
\quad \hat{y}_j \in \{\text{Authentic}, \text{Generated}\}.
\end{equation}
If the predicted label $\hat{y}_j$ matches the ground truth $y_j$, we retain the generated text $r_j$.
Otherwise, we prepend a label directive $d_j$ that explicitly specifies the target authenticity label, and regenerate the textual output under this constraint:
\begin{equation}
(r'_j, \hat{y}'_j) = f_{\text{VLM}}(x_j, p_{\text{best}}, d_j),
\quad \hat{y}'_j = y_j.
\end{equation}
The label directive is formulated as:
\begin{quote}\small
\begin{itemize}[leftmargin=1.5em]
    \item {\ttfamily If the correct label is `Generated':} \\
    {\ttfamily `This image is `Generated' (AI-generated). Explain why this image is AI-generated, not authentic.'} + $p_{\text{best}}$
    \item {\ttfamily If the correct label is `Authentic':} \\
    {\ttfamily `This image is `Authentic' (not AI-generated). Explain why this image is authentic, not AI-generated.'} + $p_{\text{best}}$
\end{itemize}
\end{quote}
This produces a regeneration dataset:
\begin{equation}
\mathcal{D}_{\text{regen}}=
\{(x_j,r_j,y_j)\mid \hat{y}_j=y_j\}
\cup
\{(x_j,r'_j,y_j)\mid \hat{y}_j\neq y_j\}.
\end{equation}
We then freeze the vision encoder and fine-tune the language module on $\mathcal{D}_{\text{regen}}$ using a standard autoregressive objective:
\begin{equation}
\mathcal{L}_{\text{text}}=
-\sum_{(x,r,y)\in \mathcal{D}_{\text{regen}}}\log p_\theta(r \mid x, y, p_{\text{best}}),
\end{equation}
where the visual encoder remains fixed and only language-side parameters are updated.
In this way, binary authenticity labels provide scalable supervision not only for the final prediction, but also for training the language-side textual outputs, without requiring human-annotated rationales.

\vspace{0.2cm}
\noindent\textbf{Scope of Textual Outputs }
Because our supervision is limited to a binary authenticity label, the generated reasoning traces are not guaranteed to be faithful to image evidence, and hallucination may occur.
We do not evaluate or claim faithfulness of these texts as verified explanations.
In UC-VLM, they are used only as an auxiliary training signal to refine language-side behavior under the same authenticity supervision, supporting more consistent outputs for detection rather than ground-truth justification.

\section{Experiments}
\noindent\textbf{Datasets and Metrics }
We evaluate \textbf{UC-VLM} on two challenging benchmarks, \textbf{GenImage} and \textbf{Chameleon}, which cover diverse real and synthetic image domains. Following the official protocols and prior works~\cite{zhu2023genimage,luo2024lare,yan2025sanity,tan2024rethinking}, we report classification accuracy (ACC) as the primary evaluation metric.

\noindent\textbullet \hspace{0.2cm} \textbf{GenImage}~\cite{zhu2023genimage} is a large-scale benchmark for AIGI detection across diverse generative models. It contains over 2.68 million images, including 1.33 million real images from ImageNet~\cite{deng2009imagenet} and 1.35 million generated images from eight models: BigGAN~\cite{brock2018largebiggan}, GLIDE~\cite{nichol2021glide}, VQDM, Stable Diffusion V1.4/V1.5~\cite{rombach2022high}, ADM~\cite{dhariwal2021diffusion}, Midjourney~\cite{Midjourney}, and Wukong~\cite{wukong}. The dataset is organized into eight subsets by source generator, enabling  per-model evaluation.

\noindent\textbullet \hspace{0.2cm} \textbf{Chameleon}~\cite{yan2025sanity} is a high-resolution test-only benchmark with 26,000 images at resolutions ranging from 720P to 4K. Real images are collected from Unsplash~\cite{uns}, while AI-generated images are collected from platforms such as ArtStation~\cite{artstation}, Civitai~\cite{civitai}, and Liblib~\cite{liblib}. The synthetic samples are mainly produced by commercial systems such as Midjourney~\cite{Midjourney} and DALL$\cdot$E 3~\cite{ramesh2022hierarchical}, or by custom LoRA~\cite{hu2022lora}-based models built on Stable Diffusion~\cite{rombach2022high}. Following the standard protocol, we report results under both ProGAN and Stable Diffusion V1.4 (SDV1.4) training settings.

\vspace{0.3cm}
\noindent\textbf{Implementation Details }
We implement UC-VLM based on Qwen2.5-VL-7B~\cite{bai2025qwen2}, initializing both the vision encoder and the language model from the official checkpoint.
All experiments are run on a single NVIDIA A100 GPU with mixed-precision (FP16) training.
Instruction optimization is performed offline using GPT-4o: starting from a base template, we generate $N{=}5$ candidates and evolve them for $M{=}10$ rounds, retaining top-$k~({k=2})$ by validation accuracy. The best instruction (\textit{Prompt-B}) is selected from candidate instructions using sampled examples from the corresponding dataset, and is then fixed for both training and inference.
For visual adaptation, we apply LoRA to the vision encoder (rank 16) and train for one epoch using SGD with learning rate $1\times10^{-4}$ and batch size 32.
For language-side adaptation, we fine-tune the language model with LoRA (rank 8) on $\mathcal{D}_{\text{regen}}$ for one epoch with learning rate $1\times10^{-4}$, while keeping the vision encoder fixed.

\begin{table}[t]
	\centering
    \renewcommand{\arraystretch}{1.1}
    \caption{\textbf{Comparisons between the proposed UC-VLM and state-of-the-art methods on the GenImage testing set.} The symbol \dag indicates our reproduction using the source code.}\vspace{-1em}
	\resizebox{\columnwidth}{!}{
		\begin{tabular}{l|l|cccccccc|c}
			\toprule[1pt]
			\multirow{2}{*}{Method} & \multirow{2}{*}{Venue} & \multicolumn{8}{c|}{Testing Subset}                                                                                           & \multirow{2}{*}{\begin{tabular}[c]{@{}c@{}}Avg\\ Accuracy (\%)\end{tabular}} \\ \cline{3-10}
			& & Midjourney    & SDV1.4        & SDV1.5        & ADM           & GLIDE         & Wukong        & VQDM          & BigGAN        &                                                                         \\ \hline
            ResNet-50~\cite{he2016deepresnet}  &  CVPR'16   & 54.9 & \textbf{99.9} & 99.7 & 53.5 & 61.9 & 98.2 & 56.6 & 52.0 & 72.1 \\
            DeiT-S~\cite{touvron2021training}& ICML'21 & 55.6 & \textbf{99.9} & 99.8 & 49.8 & 58.1 & 98.9 & 56.9 & 53.5 & 71.6 \\
            Swin-T~\cite{liu2021swin}    &  ICCV'21   & 62.1 & \textbf{99.9} & 99.8 & 49.8 & 67.6 & 99.1 & 62.3 & 57.6 & 74.8 \\
            CNNSpot~\cite{wang2020cnnspot}    &  CVPR'20  & 52.8 & 96.3 & 95.9 & 50.1 & 39.8 & 78.6 & 53.4 & 46.8 & 64.2 \\
            Spec~\cite{zhang2019detecting} & WIFS'19  & 52.0 & 99.4 & 99.2 & 49.7 & 49.8 & 94.8 & 55.6 & 49.8 & 68.8 \\
            F3Net~\cite{qian2020thinking}  & ECCV'20  & 50.1 & \textbf{99.9} & \textbf{99.9} & 49.9 & 50.0 & \textbf{99.9} & 49.9 & 49.9 & 68.7 \\
            GramNet~\cite{liu2020global}   & CVPR'20  & 54.2 & 99.2 & 99.1 & 50.3 & 54.6 & 98.9 & 50.8 & 51.7 & 69.9 \\

			DIRE~\cite{wang2023dire}         & ICCV'23              & 65.8          & 99.7          & 99.7          & 54.5          & 58.1          & 99.4          & 54.3          & 49.8          & 72.7                                                                      \\
            FatFormer\dag~\cite{liu2024forgery}    & CVPR'24              & 93.1          & 97.0          & 97.2          & 82.0          & 95.0          & 95.8          & 88.8          & 49.9          & 87.4                                                                      \\ 
            NPR~\cite{tan2024rethinking}    & CVPR'24              & 81.0          & 98.2          & 97.9          & 76.9          & 89.8          & 96.9         & 84.1          & 84.2          & 88.6                                                                      \\
            DRCT~\cite{chen2024drct}  & ICML'24              & 91.5          & 95.0          & 94.4          & 79.4          & 89.2          & 94.7          & 90.0         & 81.7          & 89.5                                                                      \\ 
            VIB-Net~\cite{zhang2025towards}  & CVPR'25              & 88.1          & 99.6          & 99.2          & 73.9          & 74.3          & 98.3          & 89.4         & -          & 88.9                                                                      \\ 
            AIDE~\cite{yan2025sanity}    & ICLR'25              & 79.4          & 99.7          & 99.8          & 78.5          & 91.8          & 98.7          & 80.3          & 66.9          & 86.9                                                                      \\ 
            DEUA~\cite{huang2025diffusion}    & ICCV'25              & 86.4          & 96.5          & 96.2          & 85.3          & 94.4          & 96.2          & 93.1          & 84.2          & 91.5                                                                      \\ 
            FIND~\cite{li2026find} & AAAI'26 & 82.9          & 89.7          & 89.8          & \textbf{87.4}          & 97.8          & 87.0          & 89.3          & 83.0          & 88.4 \\\hline
			\rowcolor{gray!15}
			UC-VLM                            & -          & \textbf{90.1}                 & 99.9          & 99.8          & 87.0          & \textbf{98.5}          & 96.9          & \textbf{98.1}          & \textbf{98.8}          & \textbf{96.1}                                                                      \\ \bottomrule
	\end{tabular}}
	\label{table1}
\end{table}

\begin{table*}[t]
	\centering
    \caption{\textbf{Comparisons between the proposed UC-VLM and state-of-the-art methods in terms of accuracy on the Chameleon dataset.}}\vspace{-1em}
\renewcommand{\arraystretch}{1.2}
\resizebox{\textwidth}{!}{
\begin{tabular}{c|cccccccccc|c}
\toprule[1pt]
\multirow{2}{*}{Training Set} & \multicolumn{11}{c}{Methods} \\ \cline{2-12}
                  & CNNSpot~\cite{wang2020cnnspot}    & FreDect~\cite{frank2020leveraging}        & Fusing~\cite{ju2022fusing}        & GramNet~\cite{liu2020globalGramNet}           & LNP~\cite{liu2022detecting}         & UniFD~\cite{ojha2023towards}        & DIRE~\cite{wang2023dire}          & Patchcraft~\cite{zhong2023rich}       & NPR~\cite{tan2024rethinking}   & AIDE~\cite{yan2025sanity} & UC-VLM                                                                      \\ \hline
ProGAN & 56.9 & 55.6 & 57.0 & 58.9 & 57.1 & 57.2 & 58.2 & 53.8 & 57.3 & 58.4 & \textbf{69.6}\\ 
SDV1.4                  & 60.1 & 56.9 & 57.1 & 61.0 & 55.6 & 55.6 & 59.7 & 56.3 & 58.1 & 62.6 & \textbf{77.9} \\
\bottomrule
\end{tabular}}
    \label{tbl_cml}
\vspace{-0.3cm}
\end{table*}

\subsection{Comparison with State-of-the-Arts}
We compare UC-VLM with recent state-of-the-art methods on two benchmarks, \textbf{GenImage} and \textbf{Chameleon}. Table~\ref{table1} reports the results on the GenImage test set. Conventional CNN-based and Transformer-based detectors~\cite{he2016deepresnet,touvron2021training,liu2021swin,wang2020cnnspot,zhang2019detecting,qian2020thinking,liu2020global} can achieve strong performance on specific generators, but their performance often degrades across different generative settings. More recent detectors such as DIRE~\cite{wang2023dire}, FatFormer~\cite{liu2024forgery}, VIB-Net~\cite{zhang2025towards}, NPR~\cite{tan2024rethinking}, DRCT~\cite{chen2024drct}, and DEUA~\cite{huang2025diffusion} further improve accuracy by introducing reconstruction cues, frequency-aware priors, or sampling-related signals.
As shown in Table~\ref{table1}, UC-VLM achieves an average accuracy of \textbf{96.1\%} on GenImage, outperforming the strongest prior result (\textbf{91.5\%} by DEUA) by \textbf{4.6\%}. UC-VLM also delivers the best or among the best results on nearly all source generators, covering both diffusion-based and GAN-based models. These results indicate that the proposed unified binary-supervised framework can effectively adapt a VLLM to AIGI detection across diverse generative settings.

We further evaluate UC-VLM on the challenging Chameleon benchmark~\cite{yan2025sanity}, which contains 26K high-resolution images spanning diverse content and generative sources. We compare against a broad set of detectors, including CNNSpot~\cite{wang2020cnnspot}, FreDect~\cite{frank2020leveraging}, Fusing~\cite{ju2022fusing}, LNP~\cite{liu2022detecting}, UniFD~\cite{ojha2023towards}, DIRE~\cite{wang2023dire}, Patchcraft~\cite{zhong2023rich}, NPR~\cite{tan2024rethinking}, and AIDE~\cite{yan2025sanity}. As shown in Table~\ref{tbl_cml}, UC-VLM achieves the best performance under both training protocols, reaching \textbf{69.6\%} under ProGAN training and \textbf{77.9\%} under SDV1.4 training. Compared with the strongest prior baseline, AIDE, UC-VLM improves accuracy by \textbf{11.2\%} under ProGAN training and \textbf{15.3\%} under SDV1.4 training. These results further support the effectiveness of UC-VLM on challenging high-quality synthetic images.

\begin{table}[t]
  \centering
  \renewcommand{\arraystretch}{1.1}
\caption{\textbf{Ablation study of UC-VLM components.}
Baseline (Visual) and Baseline (LLM) denote fine-tuning only the vision encoder or the language module, respectively, while Baseline (Visual + LLM) denotes naive joint fine-tuning of both.}
  \resizebox{0.9\columnwidth}{!}{
  \begin{tabular}{l|cc}
  \toprule[1pt]
  \textbf{Setting} & GenImage (\%)  & Chameleon (\%)\\
  \hline
  \multicolumn{3}{l}{\textit{Baseline Methods}}\\
  \hline
  Baseline (w/o fine-tuning)     & 51.9 & 55.5 \\
  Baseline (Visual)     & 76.8 & 50.9 \\
  Baseline (LLM)    & 80.3 & 71.0 \\
  Baseline (Visual + LLM)  & 77.6 & 52.0 \\
  \hline
  \multicolumn{3}{l}{\textit{Component-wise Ablations}}\\
  \hline
  Baseline (w/o fine-tuning) + Instruction Optimization  & 74.2 & 68.3 \\  
  Baseline (Visual) + Visual Adaptation     & 91.7 & 55.4 \\
  Baseline (LLM) + Label-Conditioned Regeneration  & 82.4 & 65.1 \\
  \hline
  \hline
  UC-VLM & \textbf{96.1} & \textbf{77.9}\\
  \bottomrule[1pt]
    \end{tabular}}\vspace{-1em}
    \label{Tbl:ablation}
\end{table}

\begin{figure}[t]
\centering
\includegraphics[width=0.9\columnwidth]{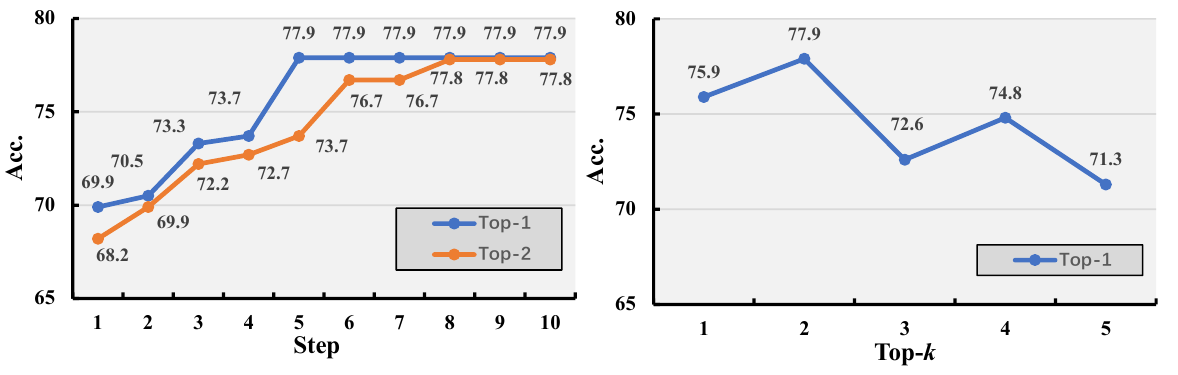}
\vspace{-1em}\caption{\label{fig:step}\textbf{Visualization of results on Chameleon under different step and Top-$k$.} According to the results, we finally set the step to 10 and the $k$ to 2.}\vspace{-1em}
\end{figure}

\begin{figure}[t]
\centering
\includegraphics[width=1.0\columnwidth]{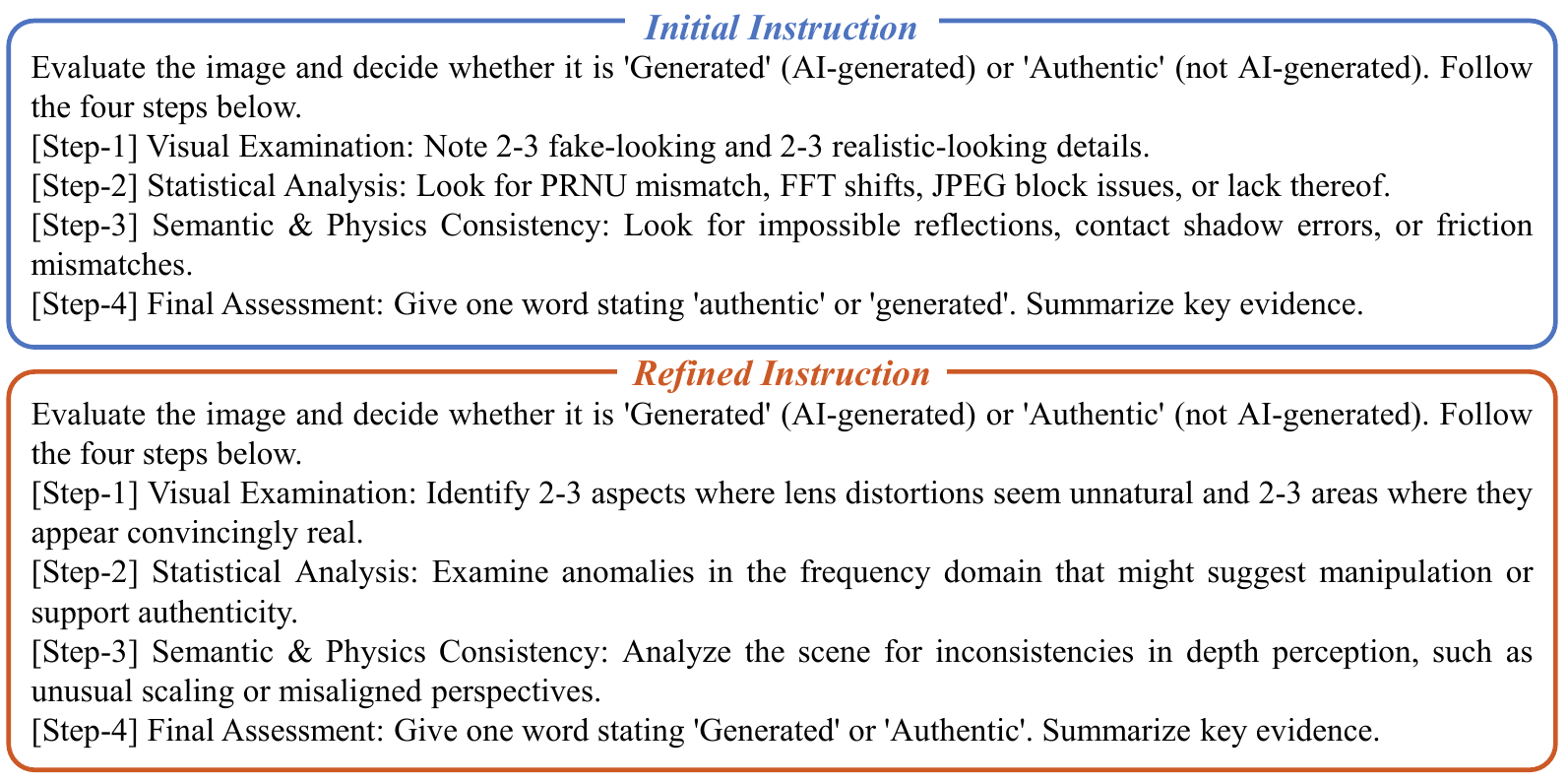}
\vspace{-2em}\caption{\label{fig_prompt}\textbf{Instruction comparison on the Chameleon dataset.} After instruction optimization, UC-VLM’s evolved prompt shifts from generic artifact descriptions to more specific visual cues, demonstrating stronger grounding in visual information.}\vspace{-1em}
\end{figure}

\begin{figure}[t]
\centering
\includegraphics[width=\columnwidth]{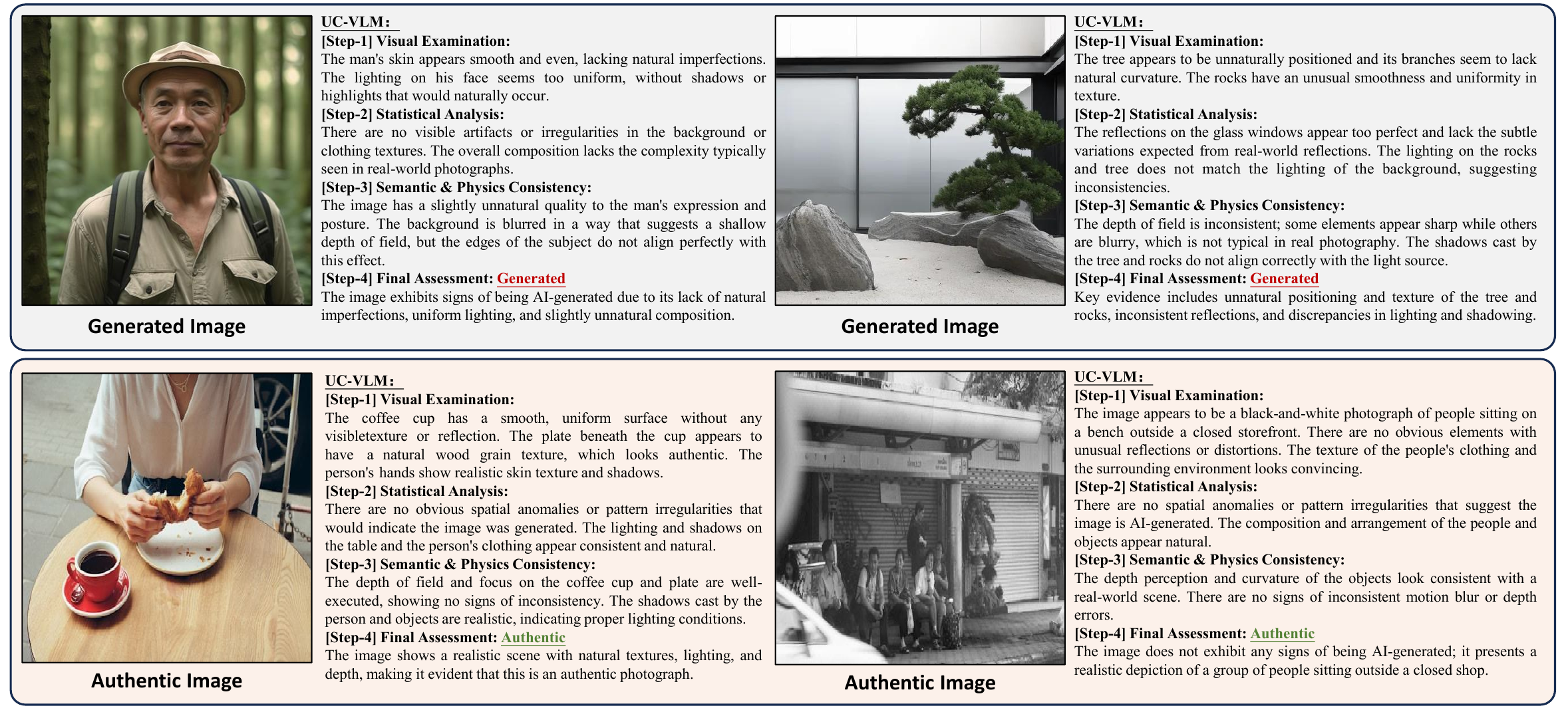}
\caption{\label{fig_vis}\textbf{Qualitative results of UC-VLM on Chameleon dataset.} UC-VLM provides structured four-step explanations for both authentic and AI-generated images, 
often describing texture irregularities, lighting inconsistencies, depth errors, and semantic-physics mismatches. These examples illustrate its qualitative behavior.}\vspace{-1em}
\end{figure}\vspace{-1em}

\begin{figure}[t]
\centering
\includegraphics[width=\columnwidth]{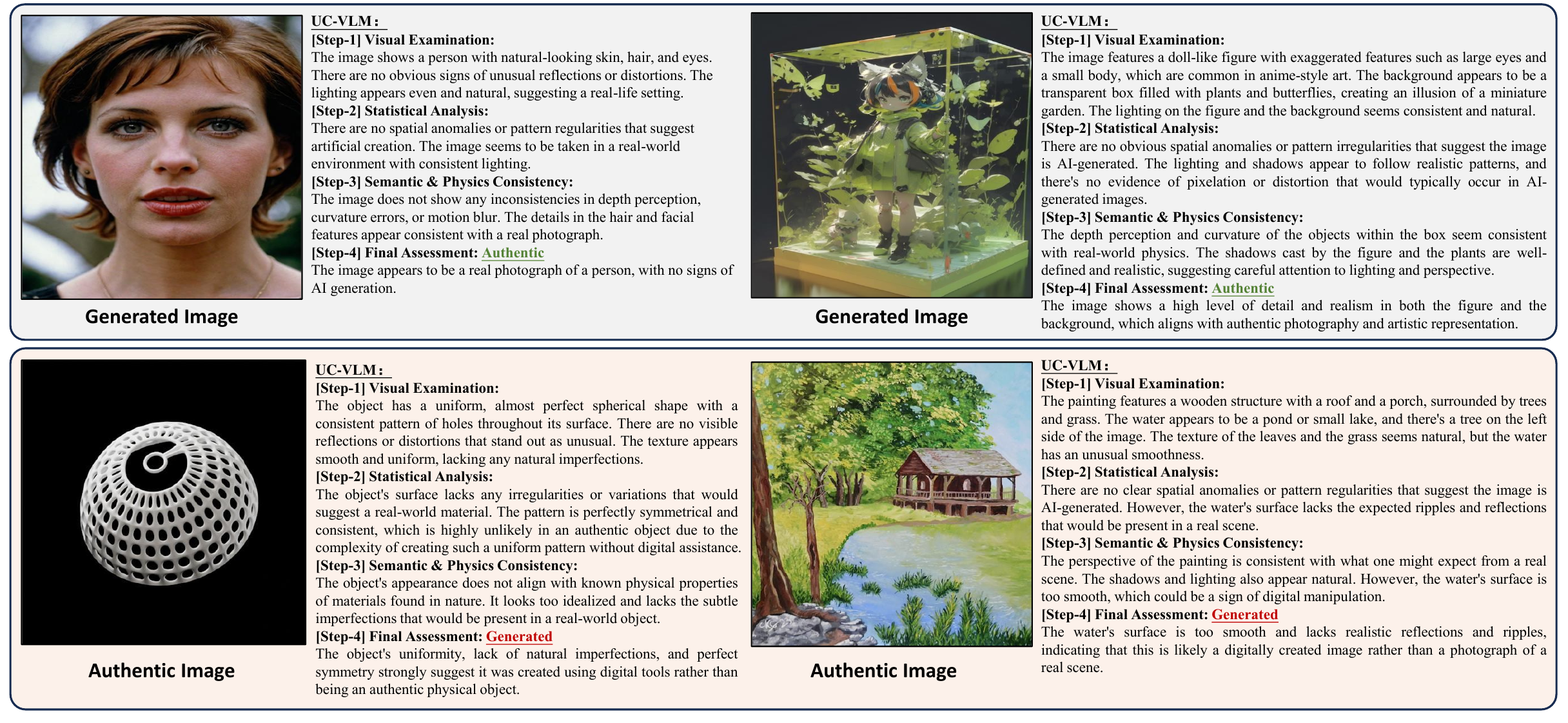}
\caption{\label{fig_fail}\textbf{Failure Cases of UC-VLM on Chameleon dataset.} UC-VLM may misclassify highly realistic generated photos as authentic when artifacts are visually subtle, and may also mistake overly pristine or heavily post-processed authentic images. It is also uncertain on artistic or stylized imagery, where coherent generated artworks can be judged as authentic while authentic paintings can be judged as generated.}\vspace{-1em}
\end{figure}

\subsection{Ablation Studies}
We conduct ablation experiments to analyze the contribution of each component in UC-VLM, as summarized in Table~\ref{Tbl:ablation}. 
Our goal is to examine how visual adaptation, instruction optimization, and label-conditioned regeneration each affect AIGI detection.
Starting from the frozen backbone (first row), Qwen2.5-VL-7B shows limited discriminative ability on both GenImage and Chameleon, indicating that a generic VLLM is not directly optimized for subtle generative artifacts. 
Fine-tuning the visual encoder alone improves performance on GenImage but brings only limited gains on Chameleon, suggesting that artifact-sensitive visual features are useful but not sufficient for robust generalization. 
In contrast, adapting the language module yields larger improvements on both datasets, highlighting the importance of language-side refinement for authenticity prediction.
Naively updating both the visual and language modules together does not produce consistent gains. This suggests that simply tuning both pathways under binary supervision is insufficient to coordinate visual and linguistic adaptation effectively.

We then evaluate each proposed component individually. 
Instruction optimization improves the frozen backbone, especially on Chameleon, confirming that VLLM predictions are sensitive to instruction formulation. 
Visual adaptation further strengthens the visual baseline by improving sensitivity to artifact-level cues. 
Label-conditioned regeneration improves the language baseline on GenImage, although its isolated effect is weaker on Chameleon.
However, combining all components yields the full UC-VLM model, which achieves the best performance on both datasets (96.1\% on GenImage and 77.9\% on Chameleon). 
These results show that the three components are complementary, and that their integration leads to more robust AIGI detection than optimizing each component in isolation.

\subsection{Discussion}
\noindent\textbf{Hyperparameters of Instruction Optimization }  
Figure~\ref{fig:step} analyzes the effect of two key hyperparameters in instruction optimization: the number of evolution rounds $M$ and the Top-$k$ selection size $k$.  
As $M$ increases, detection performance improves and then gradually saturates, indicating that iterative instruction refinement helps identify more effective formulations for authenticity assessment.

This trend also reflects the sensitivity of VLLMs to prompt formulation. Although the backbone model already contains useful visual-textual knowledge, its detection performance depends strongly on how the task is queried. This motivates automatic instruction optimization instead of relying solely on manually designed prompts.
A moderate Top-$k$ value gives the best performance. When $k$ is too small, the search tends to converge prematurely; when $k$ is too large, the instruction pool contains more redundant and lower-quality variants, which weakens the refinement process.
Overall, these results suggest that the proposed optimization process provides an effective balance between preserving strong candidate instructions and exploring diverse alternatives, enabling the model to discover more effective instruction variants for AIGI detection.

\vspace{0.2cm}
\noindent\textbf{Instruction Variation for UC-VLM }  
Figure~\ref{fig_prompt} compares the best prompts obtained before and after Instruction Optimization.  
The earlier prompt tends to enumerate a scattered set of low-level cues, such as PRNU mismatch, FFT shifts, and JPEG block effects. 
While these cues are relevant to authenticity assessment, they appear broad and are not clearly prioritized. After instruction optimization, the optimized prompt becomes more structured and places greater emphasis on cues such as frequency patterns, lens distortion abnormalities, depth inconsistency, and perspective mismatch.  
This shift suggests that adapting the vision pathway changes the visual evidence the model relies on, which in turn affects which instruction formulations are most effective.
Overall, these results indicate that Instruction Optimization moves the model from loosely collected artifact cues toward more task-oriented visual signals, providing a stronger basis for downstream detection and label-conditioned generation.

\vspace{0.2cm}
\noindent\textbf{Qualitative Results }  
Figure~\ref{fig_vis} shows qualitative examples that illustrate UC-VLM's behavior under the proposed unified learning framework.  
We emphasize that the generated text is not treated as a verified explanation. Instead, UC-VLM uses label-conditioned language-side refinement and optimized instructions as an auxiliary signal to stabilize predictions under binary supervision.
Across examples, UC-VLM identifies visual irregularities commonly observed in synthesized content and maintains reliable predictions on authentic images. Compared with the backbone model, its outputs are more stable and are more often consistent with the target authenticity label across diverse cases.

\vspace{0.2cm}
\noindent\textbf{Failure Cases}  
In Figure~\ref{fig_fail}, we show several typical failure cases of UC-VLM. First, highly realistic generated images with natural facial details, consistent lighting, and plausible geometry can be misclassified as authentic, since their synthesis artifacts are visually subtle. Second, the model shows clear uncertainty on artistic or non-photographic images: a generated digital artwork may be judged as authentic, while an authentic painting may be predicted as generated. These cases suggest that the detector has not fully learned an authenticity criterion independent of visual style and instead, it may still confuse photorealism or stylistic regularity with real/fake evidence.

\section{Conclusion}
We introduced \textbf{UC-VLM}, a unified learning framework for AI-generated image detection under binary authenticity supervision.
UC-VLM integrates visual adaptation, robustness to instruction variation, and
label-conditioned text generation within a unified multi-stage framework.
Our key novelty lies in formulating these components as one binary-supervised learning problem, so that the same authenticity label is reused to constrain both the visual pathway and the language output without requiring human-annotated rationales or manual prompt engineering.
This design strengthens sensitivity to low-level forensic cues, reduces sensitivity to prompt variation, and provides scalable supervision for textual outputs.
Our results show that unified binary-supervised learning is an effective way
to adapt VLLMs for AIGI detection.

\newpage
\section*{Acknowledgment}
This document is supported by the Ministry of Education, Singapore,
under its MOE AcRF Tier 3 Grant (MOE-MOET32022-0001).
\bibliographystyle{splncs04}
\bibliography{main}
\end{document}